\documentclass[lettersize,journal]{IEEEtran}
\usepackage{amsmath,amsfonts}
\usepackage{algorithm}
\usepackage{array}
\usepackage[caption=false,font=normalsize,labelfont=sf,textfont=sf]{subfig}
\usepackage{textcomp}
\usepackage{stfloats}
\usepackage{url}
\usepackage{verbatim}
\usepackage{graphicx}
\usepackage{cite}
\usepackage{amsthm}
\usepackage{booktabs}
\usepackage{algpseudocode}
\usepackage{makecell}
\usepackage{color}
\usepackage{caption}

\begin{document}

\title{Decorr: Environment Partitioning for Invariant Learning and OOD Generalization}

\author{Yufan~Liao, 
	    Qi~Wu, 
	    Zhaodi~Wu, 
	    and~Xing~Yan
\thanks{Yufan Liao, Zhaodi Wu, and Xing Yan are with the Institute of Statistics and Big Data, Renmin University of China, Beijing, China. (E-mail: \{liaoyf,wuzhaodianiya,xingyan\}@ruc.edu.cn.)}
\thanks{Yufan Liao and Qi Wu are with the School of Data Science, City University of Hong Kong, Kowloon Tong, Hong Kong SAR. (E-mail: yufanliao2-c@my.cityu.edu.hk, qi.wu@cityu.edu.hk.)}
\thanks{Corresponding author: Xing Yan.}}

\markboth{Journal of \LaTeX\ Class Files,~Vol.~14, No.~8, August~2021}%
{Shell \MakeLowercase{\textit{et al.}}: A Sample Article Using IEEEtran.cls for IEEE Journals}

\IEEEpubid{0000--0000/00\$00.00~\copyright~2021 IEEE}

\maketitle

\begin{abstract}
Invariant learning methods, aimed at identifying a consistent predictor across multiple environments, are gaining prominence in out-of-distribution (OOD) generalization. Yet, when environments aren't inherent in the data, practitioners must define them manually. This environment partitioning—algorithmically segmenting the training dataset into environments—crucially affects invariant learning's efficacy but remains underdiscussed. Proper environment partitioning could broaden the applicability of invariant learning and enhance its performance. In this paper, we suggest partitioning the dataset into several environments by isolating low-correlation data subsets. Through experiments with synthetic and real data, our Decorr method demonstrates superior performance in combination with invariant learning. Decorr mitigates the issue of spurious correlations, aids in identifying stable predictors, and broadens the applicability of invariant learning methods.
\end{abstract}

\begin{IEEEkeywords}
OOD generalization, IRM, environment partitioning, decorrelation.
\end{IEEEkeywords}

\section{Introduction}
\IEEEPARstart{M}{achine} learning methods have made significant strides in image classification, speech recognition, machine translation, and other domains. However, these methods typically assume that the training and testing data are independently and identically distributed (i.i.d.), an assumption that may not hold in real-world applications like autopilot, healthcare, and financial prediction. The reliance on this assumption makes using these models risky in such critical applications, as performance can drastically decline at test time, and the cost of failure is substantial \cite{shen2021towards,zhou2021domain,geirhos2020shortcut}. Invariant Risk Minimization (IRM), introduced in \cite{arjovsky2019invariant}, addresses this Out-of-Distribution (OOD) generalization issue and has garnered considerable interest. IRM's objective is to discover a data representation that ensures the optimal classifier remains consistent across all environments, thereby enhancing generalizability to new testing environments or distributions. This approach has proven successful in various scenarios and datasets. Building on IRM's principles, several other invariant learning methods \cite{rosenfeld2020risks,ahmed2020systematic,ahuja2020invariant,lu2021nonlinear,chattopadhyay2020learning} have emerged, achieving promising OOD performance.

To deploy these invariant learning methods, establishing an environment partition of the training set is necessary. Existing approaches often rely on data sources or metadata to determine this partition. However, a natural partition may not always exist or can be difficult to identify, rendering these methods unsuitable for many datasets \cite{sohoni2020no}. For instance, in the Colored MNIST (CMNIST) synthetic dataset, the environments are defined with the correlations $\text{corr}(\text{color}, \text{number}) = 0.8$ and $0.9$ respectively, and the belonging of each image to a specific environment is known during training. A more realistic scenario is one where such environment information is unknown. Even when a natural environment partition exists, it's worth questioning whether it is optimal for developing a model that generalizes well, considering the myriad ways the data could be segmented into different environments.

\IEEEpubidadjcol

Several studies have addressed the challenge of environment partitioning. Creager et al. \cite{creager2021environment} introduced Environment Inference for Invariant Learning (EIIL), which employs a reference classifier $\Phi$ trained using ERM to identify partitions that maximally violate the invariance principle, thereby maximizing the IRMv1 penalty on $\Phi$. Similarly, Just Train Twice (JTT) \cite{liu2021just} involves training a reference model first, followed by a second model that upweights training examples misclassified by the initial model. However, the effectiveness of these two-step, mistake-exploiting methods depends significantly on the performance of the reference model \cite{nam2020learning,dagaev2021too}. Another straightforward approach to partitioning is clustering. Works by Matsuura et al. \cite{matsuura2020domain}, Sohoni et al. \cite{sohoni2020no}, and Thopalli et al. \cite{thopalli2021improving} have utilized conventional clustering techniques like $k$-means to divide the dataset based on feature space, while Liu et al. \cite{liu2021heterogeneous} sought to maximize the diversity of the output distribution $P(Y|\Phi(X))$ through clustering.

Although training and testing sets are not i.i.d. in the OOD setting, they should share certain common properties that aid in generalization. A widely recognized OOD assumption addressing covariate shift posits that $P_{\text{train}}(Y|X) = P_{\text{test}}(Y|X)$ \cite{shen2021towards}, suggesting that the outcome distribution remains consistent across training and testing sets given $X$. Building on this premise, it follows that the environment label $e_X$, whether based on $X$ or features extracted as $\Phi(X)$, should be independent of the outcome $Y$. This foundational assumption enables us to address discrepancies between training and testing environments effectively.

However, most environment partitioning methods utilize the outcome $Y$ for segmentation. This approach is less harmful in scenarios with a high signal-to-noise ratio, such as image data, where $X$ inherently encompasses most information about $Y$. However, in the presence of noisy data, such as tabular data, these methods may mistakenly assign similar or identical features into different environments due to variations in $Y$. For instance, two data points with the same $X$ might end up in separate environments solely because the error components (irreducible and purely stochastic) in their $Y$ values differ. This leads to disparate conditional distributions of the outcome across environments, violating the covariate shift assumption. The efficacy of these methods in high-noise contexts remains largely unexplored. While $k$-means clustering does not depend on $Y$, relying solely on the feature space, it is not supported by a clear interpretation or theoretical justification for its application in this context.

This paper explores an environment partitioning method tailored for high-noise data to enhance the OOD generalization performance of IRM. We observe that models trained on datasets with uncorrelated features generally perform well against correlation shifts. IRM specifically seeks to develop an invariant predictor that excels across various environments. Inspired by these observations, we introduce Decorr, a method designed to identify subsets of features with low correlation for environment partitioning. Decorr is computationally efficient and independent of the outcome $Y$. Through experiments with both synthetic and real data, we demonstrate that Decorr, in conjunction with IRM, consistently outperforms in OOD scenarios we established, whereas some existing partitioning methods paired with IRM yield poor, sometimes worse-than-ERM, results.

To summarize, in this paper, our contributions are:
\begin{enumerate}
	\item We introduce Decorr, a method that manually partitions the dataset into environments to enhance OOD generalization in combination with invariant learning. Decorr identifies subsets characterized by features with low correlations, which mitigates the issue of spurious correlations and aids in identifying stable predictors.
	\item We demonstrate through simulation studies that Decorr-based invariant learning can achieve good OOD generalization even under model misspecification.
	\item The proposed Decorr method demonstrates superior performance across a diverse range of datasets, including both tabular and image types.
	\item We broaden the applicability of invariant learning methods (IRM, REx, etc.) when natural environment partitions are unknown. We improve the performance of invariant learning when the natural environment partitions are suboptimal.
\end{enumerate}

\section{Background}

\subsection{Invariant Risk Minimization}
IRM \cite{arjovsky2019invariant} works with datasets $D_e = \{(x_i^e, y_i^e)\}$ sourced from multiple training environments $e \in \mathcal{E}_{tr}$, aiming to develop a model that excels across an extensive range of environments $\mathcal{E}_{all}$, where $\mathcal{E}_{tr} \subset \mathcal{E}_{all}$. The objective is to minimize the worst-case risk $R^{ood}(f) = \max\limits_{e \in \mathcal{E}_{all}}R^e(f)$, with $R^e(f) = \mathbb{E}_e[l(f(x),y)]$ representing the risk within environment $e$. Specifically, IRM seeks to identify a data representation $\Phi$ and a classifier $w$ that remains optimal across all training environments $\mathcal{E}_{tr}$ when using the representation $\Phi$. This challenge is framed as a constrained optimization problem:
\begin{equation}
	\begin{split}
		&  \min\limits_{w,\Phi} \sum\limits_{e \in \mathcal{E}_{tr}} R^e(w \circ \Phi), \\
		\text{subject to } & w \in \mathop{\arg\min}\limits_{\bar{w}} R^e(\bar{w} \circ \Phi), \text{ for all } e \in \mathcal{E}_{tr}. \\
	\end{split}
	\label{IRM}
\end{equation}
To make the problem solvable, the practical version IRMv1 is expressed as 
\begin{equation}
	\min\limits_{f} \sum\limits_{e \in \mathcal{E}_{tr}} R^e( f) + \lambda  ||\nabla_{w|w=1} R^e(w \cdot f) ||^2,
\end{equation}
where $f$ indicates the entire invariant predictor, and $w = 1$ is a fixed dummy scalar. The gradient norm penalty can be interpreted as the invariance of the predictor $f$.

Another approach to invariant learning is Risk Extrapolation (REx) \cite{krueger2021out}, which aims to reduce training risks while increasing the similarity of training risks across environments. Variance-REx (V-REx) adds a penalty term—the variance of training losses across all training environments—to the traditional empirical risk minimization. It has been shown that this method can perform robustly in the presence of covariate shift.

\subsection{Environment Partitioning Methods}
To our best knowledge, literature predominantly features two types of partitioning methods. Clustering methods for environment partitioning are discussed in \cite{matsuura2020domain,sohoni2020no,thopalli2021improving}. The general approach involves extracting features from the data and then clustering the samples based on these features into multiple groups, with all proposed methods employing $k$-means for clustering. Conversely, EIIL \cite{creager2021environment} introduces an adversarial approach that partitions data into two environments designed to maximize the IRM penalty using an ERM model. Specifically, the environment inference step seeks to optimize a probability distribution $\mathbf{q}_i(e') := q(e'|x_i,y_i)$ to enhance the IRMv1 regularizer $C^{EI}(\Phi,\mathbf{q}) = ||\nabla_{w|w=1} \tilde{R}^e(w \circ \Phi, \mathbf{q}) ||$, where $\tilde{R}^e$ represents the $\mathbf{q}$-weighted risk. After identifying the optimal $\mathbf{q}^*$, environments are assigned by placing data into one environment based on $\mathbf{q}^*$, with the rest allocated to another environment.

\subsection{Some Other Related Works}

\subsubsection{Feature Decorrelation}
The use of correlation for feature selection is well-established in machine learning \cite{hall1999correlation,yu2003feature,hall2000correlation,blessie2012sigmis}. More recently, the decorrelation method has been adapted for stable learning and OOD generalization. \cite{zhang2021deep} suggested decorrelating features by learning weights for training samples. \cite{shen2020stable} initially clustered variables based on the stability of their correlations, then proceeded to decorrelate pairs of variables from different clusters. \cite{kuang2020stable} focused on simultaneously optimizing regression coefficients and sample weights to manage correlation. However, to our knowledge, discussions on using decorrelation for dataset partitioning to enhance invariant learning are lacking, which is the main focus of the subsequent sections in our study.

\subsubsection{Covariate Shift}
Covariate shift has been a significant challenge in machine learning long before the advent of OOD generalization and invariant learning. Earlier research on covariate shift focused on adaptively training a predictor using the training dataset, sometimes incorporating an unlabeled testing dataset or known test-train density ratios, but without employing multiple training sets or environment partitions. Under such conditions, Importance Weighting (IW) has proven to be an effective strategy for addressing covariate shift \cite{sugiyama2007covariate,sugiyama2007direct,sugiyama2008direct,bickel2009discriminative,fang2020rethinking,fang2024generalizing}. In contrast, the contexts of OOD generalization and invariant learning typically require environment partitions and do not involve any pre-knowledge of the testing set. For further insights, theories, and methodologies related to covariate shift in machine learning, please refer to \cite{sugiyama2012machine,quinonero2022dataset}.

\section{The Proposed Method}

The IRM objective outlined in Eqn. \eqref{IRM} seeks to minimize risk across a set of environments $\mathcal{E}_{tr}$, while imposing an invariant constraint on the weights $w$. IRM operates under the assumption that the environment partition is pre-established. However, in real-world scenarios, the data can be divided into environments in numerous ways. For instance, such partitioning could be based on personal characteristics like gender, age, or education level when predicting income from personal data, or it might depend on the timing of data collection. These partitioning strategies may be subject to scrutiny. Often in practice, we are either presented with a single training dataset without a clear environment partition or handed a potentially inadequate predetermined partition. Consequently, it becomes essential to manually select an environment partition that is effective for invariant learning methods to maximize the OOD performance of the model.

In this section, we explore how to identify a subset of data $\tilde{X} \subset X$ characterized by low correlation, making it ideal for IRM learning. Given a data matrix $X_{n\times p}$, we denote its correlation matrix by $R_{X} = (r_{ij})_{p\times p}$. The deviation of $R_X$ from the identity matrix $I$ is assessed using the squared Frobenius distance $d^2(R_X,I) = \sum_{i,j=1}^{p} (r_{ij} - \delta_{ij})^2$, which serves as a measurement of how uncorrelated $X$ is. The formulation of our goal is as follows:
\begin{equation}
	\min \limits_{\tilde{X} \subset {X}} d^2(R_{\tilde{X}},I),
	\label{original}
\end{equation}
subject to some constraints on the size of $\tilde{X}$. Given the large feasible set and the complexity of the optimization, we approach it by minimizing a softer alternative that transforms the subset selection into an optimization of sample weights. Using a weight vector $w_{n \times 1}$ for the observations in $X$, the weighted correlation matrix $R_X^w = (r_{ij}^w)$ is computed as described in \cite{Costa2011}. The correlation minimization problem can thus be reformulated as
\begin{equation}
	\min \limits_{w \in [0,1]^n} d^2(R_X^w,I),\quad\text{with constraints on } w,
	\label{target}
\end{equation}
which is amenable to optimization. Here, the $i$-th element of $w$ represents the probability that the $i$-th data point is included in the new environment $\tilde{X}$ from the entire set $X$. Restricting $w$ to $\{0,1\}^n$ causes (\ref{target}) to revert to the original hard problem (\ref{original}).

The specified target poses convergence challenges without constraints on $w$. To address this, we propose two restrictions. First, considering $k$ as the number of desired environments, we limit the mean of the weights with $\frac{1}{n}1^\top w = \frac{1}{k}$. This constraint prevents the predominance of excessively small values in $w$, which could lead to a small sample size and high variance within the partitioned environments. We enforce this by incorporating a penalty term $\lambda (\frac{1}{n}1^\top w - \frac{1}{k})^2$ into the objective. Additionally, we constrain $w_i \in [p_0,1]$—where $p_0$ is a minimal value near zero—rather than $w_i \in [0,1]$. This adjustment not only facilitates the optimization's convergence but also guarantees that all data points in the training set are eligible for inclusion in the partitioned set, allowing the model trained on this set to adapt to the entire distribution of observations rather than a confined segment. This approach also balances the trade-off between diversity shift and correlation shift \cite{ye2021ood}.


To partition the training dataset into environments such that $\cup_{j=1}^k {X_j} = X$, we iteratively optimize the objective specified in Eqn. \eqref{target} with respect to the residual sample set, selecting samples to establish each new environment sequentially. We detail this procedure in Algorithm \ref{algo1}. Once the environments are delineated, we employ IRM as the learning strategy to enhance OOD generalization.
\begin{algorithm}[t]
	\textbf{Input:} training set $X = \{x_i\}_{i=1}^{|X|}$, number of desired environments $k$, restriction parameter $p_0$, learning rate $\alpha$, number of epochs $T$, and $\lambda$ (suggested to be $100$) \\
	\textbf{Output:} the partitioned environments $\mathop{\cup}\limits_{j=1}^k {X_j} = X$ \\
	\textbf{Initialization:} the residual set $X_r = X$
	\begin{algorithmic}[1]
	\For{$j = 1,2,\dots,k-1$}
		\State Re-denote $X_r=\{x_i\}_{i=1}^{|X_r|}$, and randomly initialize $w_i \sim\text{Uniform}[p_0,1], ~i =1,2,\dots,|X_r|$
		\State Let $t = 0$
		\While{not converged and $t < T$}
			\State $L(w) = d^2(R^w_{X_r},I) + \lambda (\frac{1}{|X_r|}1^\top w - \frac{1}{k-j+1})^2$
			\State $w \leftarrow w-\alpha \frac{\partial{L(w)}}{\partial{w}}$
			\State $t \leftarrow t+1$
		\EndWhile
		\State Let $x_i \in X_j$ with probability $w_i$, for  $i=1,2,\dots,|X_r|$
		\State $X_r \leftarrow X_r - X_j$
	\EndFor
	\State $X_k \leftarrow X_r$
	\end{algorithmic}
	\caption{The Decorr Algorithm}
	\label{algo1}
\end{algorithm}

\begin{figure*}[t]
	\centering
	\begin{minipage}[t]{0.3\textwidth}
		\centering
		\includegraphics[width=\textwidth]{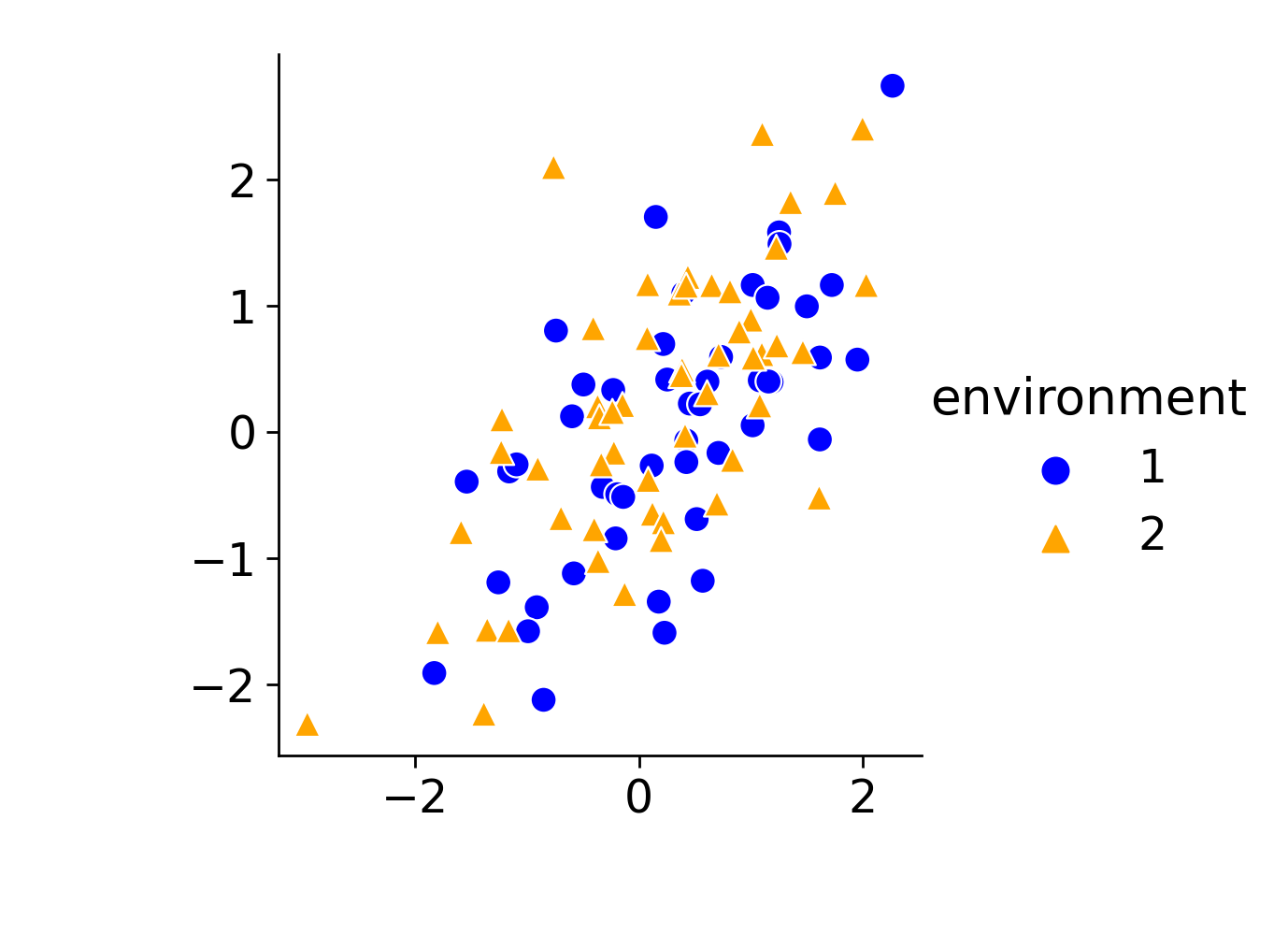}
		\vspace{-2.5em}
		\caption*{$\qquad$ EIIL}
	\end{minipage}
	\begin{minipage}[t]{0.3\textwidth}
		\centering
		\includegraphics[width=\textwidth]{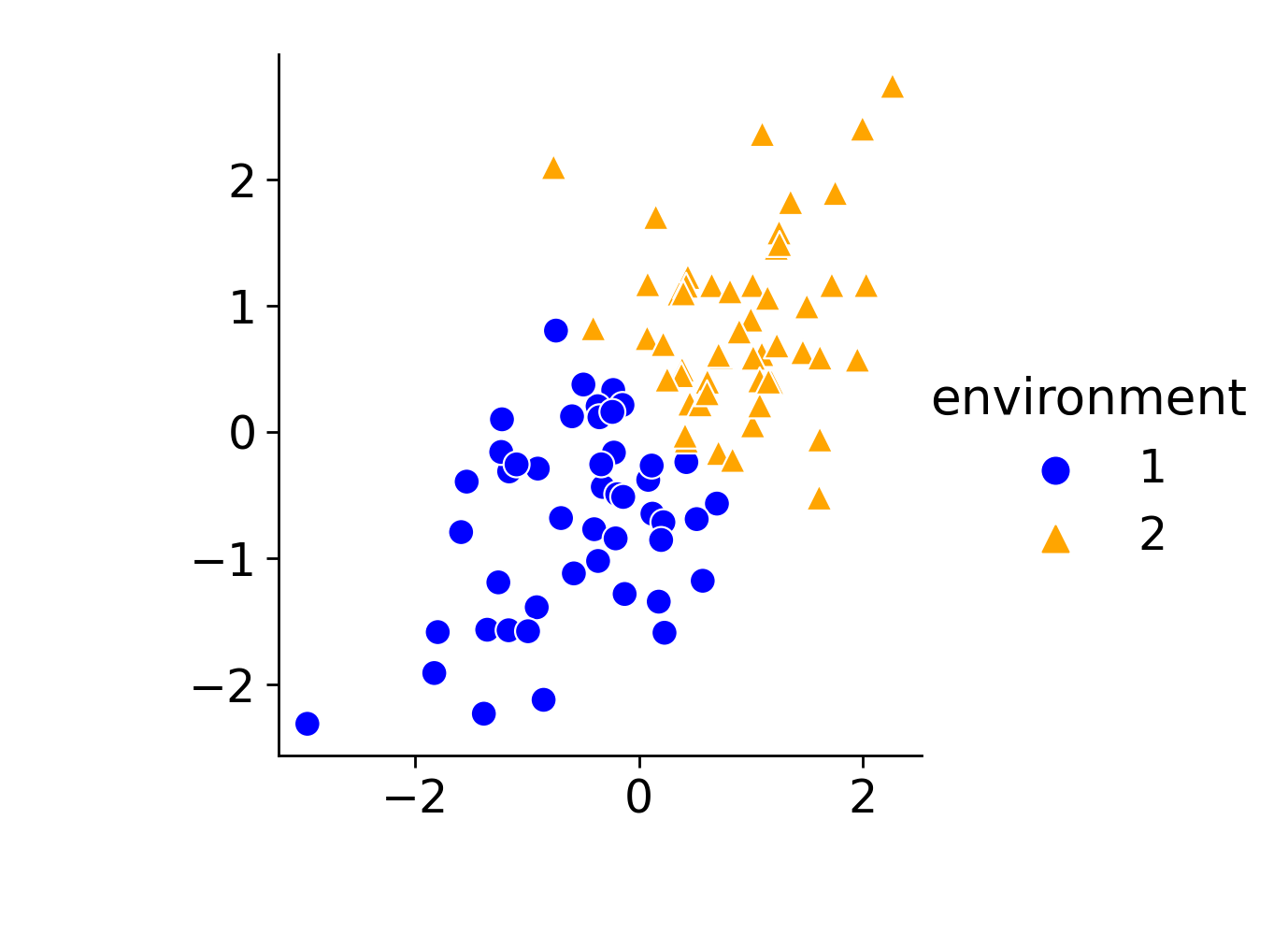}
		\vspace{-2.5em}
		\caption*{$\qquad$ $k$-means}
	\end{minipage}
	\begin{minipage}[t]{0.3\textwidth}
		\centering
		\includegraphics[width=\textwidth]{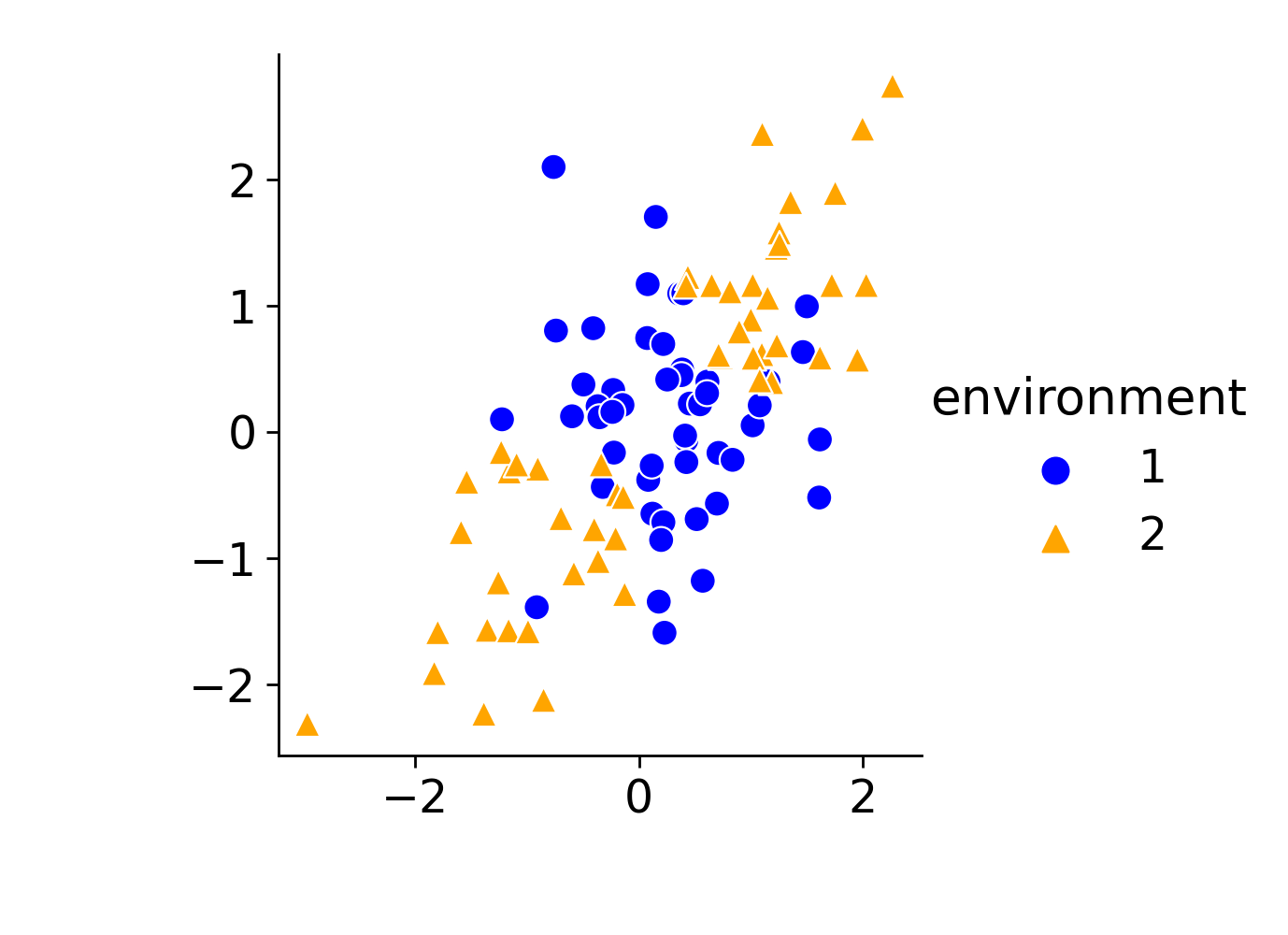}
		\vspace{-2.5em}
		\caption*{$\qquad$ Decorr}
	\end{minipage} 
	\caption{The resulting partitions on a toy dataset. The EIIL partition shows no clear patterns, suggesting a strong dependence on the label $y$ and deviating from expected environmental separations. Decorr divides the dataset into environments characterized by distinct covariate relationships: one positively correlated (triangle) and one almost uncorrelated (circle). While $k$-means also bifurcates the data spatially, the divisions it creates feature similar covariate properties with only a mean shift.}
	\label{partition}
\end{figure*}

To explore the impact of various environment partitioning strategies, we applied EIIL, $k$-means, and Decorr to a two-dimensional toy dataset where $x_0$ and $x_1$ exhibit positive correlation, and $y$ is the sum of $x_0$ and an error term. The resulting partitions are illustrated in Fig. \ref{partition}. The EIIL partition shows no clear patterns, suggesting a strong dependence on the label $y$ and deviating from expected environmental separations. Both $k$-means and Decorr reveal spatial characteristics. Decorr divides the dataset into environments characterized by distinct covariate relationships: one positively correlated (triangle) and one almost uncorrelated (circle). While $k$-means also bifurcates the data spatially, the divisions it creates feature similar covariate properties with only a mean shift, potentially diminishing its utility for IRM applications.

\subsection{OOD Generalization under Model Misspecification}

In addition to our earlier discussions, we design an experiment to demonstrate that our method can achieve OOD generalization even when the models are misspecified. We adhere to most data-generating configurations and settings as described in \cite{kuang2020stable}. We denote $S \in \mathbb{R}^{p_s\times 1}$ as the invariant features and $V \in \mathbb{R}^{p_v\times 1}$ as the spurious features, and $n$ is the sample size. The data are generated according to the model outlined below:
\begin{equation}
	\begin{split}
		& S_{j} \mathop{\sim}\limits^{\text{i.i.d}} \mathcal{N}(0,1),\quad j=1,\dots,p_s, \\
		& V_{k} \mathop{\sim}\limits^{\text{i.i.d}} \mathcal{N}(0,1),\quad k=1,\dots,p_v, \\
		& Y = S^\top \beta + f(S) + \varepsilon,
	\end{split}
\end{equation}
where  $\varepsilon \sim \mathcal{N}(0,1)$ and $f(S)$ is a nonlinear term. 
We define $f(S) = S_{1}S_{2}S_{3}$ or $f(S) = e^{S_{1}S_{2}S_{3}}$, even though $p_s$ may exceed 3. The coefficient vector $\beta = [\frac{1}{3}, -\frac{2}{3}, 1, -\frac{1}{3}, \frac{2}{3}, -1, \dots]^\top$ cycles through these six values, truncated to fit $p_s$ dimensions. Here, a straightforward $p_s = p_v = 5$ is set by us.




To create training and testing sets with selection bias, we first produce original samples using the aforementioned method, then allocate each sample to the set based on the probability $P = |r|^{-5|S^\top\beta+f(S)-\text{sign}(r)V_{1}|}$. A $r>1$ results in a positive correlation between $V$ and $Y$, while $r<-1$ leads to a negative correlation. Nonetheless, the generation of $Y$ is independent of  $V$, thereby introducing a spurious correlation through selection bias. In our experiment, $r_{\text{train}}=2$ with 20,000 samples in the training set, and $r_{\text{test}}=-2$ with 10,000 samples in the testing set.

We employ linear regression of $Y$ against $(S,V)$ as the base model to fit the synthetic data. 
We compare Decorr against other environment partitioning strategies (pure random and $k$-means), combined with two different invariant learning algorithms (IRM and V-REx \cite{krueger2021out}). Additionally, we include ERM (Empirical Risk Minimization) and Decorr+ERM, the latter of which applies ERM to the first environment divided by Decorr, as baselines. To assess each model's generalization capability, we use $\frac{1}{p_v}||\hat{\beta_v}||_1$, the $L_1$ norm of the coefficient on $V$, which should ideally be zero for an optimal model, and the Mean Squared Error (MSE) in the testing set. The experimental results for two types of $f(S)$ are presented in Table \ref{bias1} and \ref{bias2}, respectively.  The results indicate that Decorr achieves superior performance on environment partitioning relative to other methods, meanwhile underscoring the importance of using invariant learning approaches, as evidenced by the baseline performance of ERM and Decorr+ERM. The $L_1$ norm of the coefficient on $V$ and the MSE for the testing set are significantly reduced when using Decorr+IRM or Decorr+REx.

\begin{table}[t]
	\centering
	\caption{Selection bias experiment with $f(S) = S_{1}S_{2}S_{3}$. We compare Decorr against random and $k$-means, combined with IRM and V-REx. ERM and Decorr+ERM are included too. The two evaluation metrics are significantly reduced when using Decorr+IRM or Decorr+REx, underscoring both the superiority of Decorr and the importance of invariant learning.}
	\label{bias1}
	\begin{tabular}{c|c|c}
		\toprule
		Method & $\frac{1}{p_v}||\hat{\beta_v}||_1$ & Testing Set MSE \\
		\midrule
		ERM & 0.68 (0.05) & 2.24 (0.17)\\
		Decorr + ERM & 0.64 (0.06) &  2.11 (0.19) \\
		$k$-means + IRM & 0.52 (0.09) & 2.34 (0.18)\\
		Random + IRM & 0.73 (0.20) & 2.36 (0.24)\\
		\textbf{Decorr + IRM} & \textbf{0.30 (0.22)} & \textbf{1.85 (0.35)}\\
		$k$-means + REx & 0.67 (0.15) & 2.31 (0.33)\\
		Random + REx & 0.80 (0.10) & 2.34 (0.28)\\
		\textbf{Decorr + REx} & 0.65 (0.20) & \textbf{1.83 (0.12)} \\    
		\bottomrule
	\end{tabular}
\end{table}

\begin{table}[t]
	\centering
	\caption{Selection bias experiment with $f(S) = \exp(S_{1}S_{2}S_{3})$. We compare Decorr against random and $k$-means, combined with IRM and V-REx. ERM and Decorr+ERM are included too. The two evaluation metrics are significantly reduced when using Decorr+IRM or Decorr+REx, underscoring both the superiority of Decorr and the importance of invariant learning.}
	\label{bias2}
	\begin{tabular}{c|c|c}
		\toprule
		Method & $\frac{1}{p_v}||\hat{\beta_v}||_1$ & Testing Set MSE \\
		\midrule
		ERM & 0.77 (0.03) & 2.64 (0.16) \\
		Decorr + ERM &  0.78 (0.08) & 2.60 (0.24) \\
		$k$-means + IRM & 0.58 (0.16) & 2.00 (0.15)\\
		Random + IRM & 0.70 (0.25) & 2.12 (0.50)\\
		\textbf{Decorr + IRM} & \textbf{0.39 (0.25)} & 2.12 (0.36)\\
		$k$-means + REx & 0.72 (0.10) & 2.41 (0.20)\\
		Random + REx & 0.83 (0.10) & 2.58 (0.19)\\
		\textbf{Decorr + REx} & \textbf{0.38 (0.18)} & \textbf{1.64 (0.08)} \\    
		\bottomrule
	\end{tabular}
\end{table}

\section{Experiments}

We assess four distinct environment partitioning methods: pure random, EIIL \cite{creager2021environment}, $k$-means, and Decorr, along with the complete procedure of Heterogeneous Risk Minimization (HRM, \cite{liu2021heterogeneous}) in some experiments. Additionally, we deploy the original IRM (and V-REx \cite{krueger2021out} exclusively in real data experiments) where environments are pre-defined, and also employ ERM for comparative analysis. We utilize either the original or widely used implementations of IRM\footnote{https://github.com/facebookresearch/InvarianceUnitTests}, EIIL\footnote{https://github.com/ecreager/eiil}, and HRM\footnote{https://github.com/LJSthu/HRM}.

Across all experiments, we set parameters $p_0 = 0.1$, $\alpha = 0.1$, $T = 5000$, and $\lambda = 100$ for Decorr, adjusting $k$ according to the specific requirements of each scenario. For instance, in synthetic data experiments, $k$ corresponds to the actual number of environments in the training data, ranging from 2 to 8, though in most cases, $k$ is either 2 or 3. Similarly, in related studies employing methods like EIIL, HRM, or $k$-means, the number of training environments typically ranges between 2 and 3.

Although this paper primarily does not target high signal-to-noise ratio data, we showcase the adaptability of Decorr by applying it to two image datasets, CMNIST and Waterbirds, demonstrating its effectiveness across various dataset types. Feature extraction is conducted using a neural network before proceeding with the decorrelation process. Decorr is applied to these extracted low-dimensional features ($k$-means is applied similarly). After partitioning the dataset, we continue to employ the original raw image data for IRM training.

\subsection{Synthetic Data Experiments}

\subsubsection{A Toy Example}
Following \cite{aubin2021linear}, we consider a toy $d$-dimensional example generated by 
\begin{equation}
	\begin{split}
		x_1 &\sim \mathcal{N}(0,(\sigma^e)^2I), \\
		\tilde{y} &\sim \mathcal{N}(W_{yx}x_1,(\sigma^e)^2I), \\
		x_2 &\sim \mathcal{N}(W_{xy}\tilde{y},I),    
		\label{IRMEx}
	\end{split}
\end{equation}
where $x_1,~\tilde{y},~x_2 \in \mathbb{R}^{d}$. The task is to predict $y = 1_d^\top \tilde{y}$ given $x = (x_1,x_2)$. We can write $W_{xy}\tilde{y} \sim \mathcal{N}(x_2, I) $ if we ignore the causal mechanics, hence using $x_2$ to predict $y$ has a fixed noise. Therefore, if $\sigma^e$ is small, the model will rely more on $x_1$ due to the lower noise in predicting $y$ using $x_1$. Conversely, if $\sigma^e$ is large, the model will rely more on $x_2$. For simplicity, we set $W_{xy} = W_{yx} = I$ and $\sigma^e = 0.1, ~1.5, ~2$ for three different environments. In each environment $e$, 1,000 samples are generated. The goal is for the model to learn the true causal relationship, $y = 1^\top_d x_1 + \varepsilon$, rather than the spurious correlation between $x_2$ and $y$. Therefore, we assess the effectiveness of different environment partitioning strategies by calculating the Mean Squared Error (MSE) between the linear regression coefficients and the ground-truth coefficients $\beta^* = (1_d,0_d)$.

Data from original environments are combined, followed by the application of environment partitioning strategies to achieve the same number of environments, and then the application of IRM. We consider scenarios featuring varying numbers of original environments (2 or 3, corresponding to the first two $\sigma^e$ or all three) and different data dimensions $d=2,~5,~10,~20$. The penalty weight in IRM is set as $\lambda = 10$. Results, derived from 10 trials to calculate the mean and the standard deviation, are displayed in Table \ref{IRMexample}. In most instances, Decorr yields coefficients that are closest to $\beta^*$. Original IRM, when true environments are known (as indicated in the last row), reliably generates the true coefficients.

\begin{table*}[t]
	\caption{A toy example generated by Eqn. \eqref{IRMEx}. Our Decorr achieves the lowest MSE between linear regression coefficients and the ground-truth coefficients $\beta^* = (1_d,0_d)$, in various scenarios featuring varying numbers of original/partitioned environments and different data dimensions. Standard deviations over 10 trials are included.}
	\label{IRMexample}
	\begin{center}
			\begin{tabular}{ccccccccc} 
				\toprule
				\multicolumn{1}{c}{}  &\multicolumn{4}{c}{Number of Environments $=2$} & \multicolumn{4}{c}{Number of Environments $=3$}  \\
				\cmidrule(r){2-5} \cmidrule(r){6-9} 
				Method & $d=2$ & $d=5$ & $d=10$ & $d=20$ & $d=2$ & $d=5$ & $d=10$ & $d=20$\\
				\midrule
				ERM  & 0.28 (0.01) & 0.28 (0.02)&0.29 (0.02) & 0.28 (0.02)&0.46 (0.01)&0.46 (0.01)&0.46 (0.02)&0.46 (0.02)\\ 
				Random + IRM & 0.28 (0.06) & 0.27 (0.04) &0.27 (0.02)&0.29 (0.02)&0.45 (0.06)&0.49 (0.10)&0.49 (0.04)&0.46 (0.04) \\
				EIIL  & 0.29 (0.05) & 0.35 (0.06)&0.36 (0.03) &0.38 (0.04)&\textbf{0.17 (0.08)}&0.31 (0.08)&0.34 (0.03)&0.37 (0.04)\\
				$k$-means + IRM & 0.21 (0.05) &  0.27 (0.04)&0.27 (0.03) &0.28 (0.02)&0.33 (0.08)&0.33 (0.10)&0.38 (0.07)&0.35 (0.04)\\
				HRM & 0.38 (0.17) & 0.43 (0.06) & 0.44 (0.04) & 0.47 (0.02) & 0.50 (0.00) & 0.50 (0.00) & 0.50 (0.00) & 0.49 (0.01)\\
				\textbf{Decorr + IRM} & \textbf{0.13 (0.03)} & \textbf{0.16 (0.09)}& \textbf{0.16 (0.06)} & \textbf{0.09 (0.02)}&0.25 (0.02)&\textbf{0.29 (0.04)}&\textbf{0.26 (0.05)}&\textbf{0.25 (0.03)}\\
				
				\midrule
				IRM (Oracle)  & 0.02 (0.00) & 0.02 (0.01) & 0.02 (0.00)&0.02 (0.00) &0.07 (0.03)&0.04 (0.01)&0.04 (0.02)&0.02 (0.00)  \\
				\bottomrule
			\end{tabular}
		\end{center}
	\end{table*}

	\subsubsection{Risks of IRM}
	Next, we consider another example in \cite{rosenfeld2020risks}, which is generated by
		\begin{equation}
			y = \left\{
			\begin{aligned}
				&    1,\quad\text{with probability }\eta, \\
				&    -1,\quad\text{otherwise}, \\
			\end{aligned}
			\right .
			\label{risk1}
		\end{equation}
		and
		\begin{equation}
			\begin{split}
				z_c &\sim \mathcal{N}(y \cdot \mu_c, \sigma_c^2I),\\
				z_e &\sim \mathcal{N}(y \cdot \mu_e, \sigma_e^2I),\\
				x &= f(z_c,z_e).
			\end{split}
			\label{risk2}
		\end{equation}
	Assume data are drawn from $E$ training environments $\mathcal{E} = \{e_1,e_2,\dots,e_E\}$. For a given environment $e$, a data point is obtained by first randomly sampling a label $y$, then sampling invariant features $z_c \in \mathbb{R}^{d_c}$ and environmental features $z_e \in \mathbb{R}^{d_e}$. Finally, the observation $x$ is generated.
	
	We let $E$ vary from 2 to 8, and for each case, we do training and testing 10 times to average the results. In each time, we set $\eta = 0.5$, $\mu_c = Z_1 + 0.5\text{sign}(Z_1)$, $\mu_e = 1.5Z_2+Z_e$, $\sigma_c^2 = 2$, $\sigma_e^2 = 0.1$, where $Z_1\in\mathbb{R}^3$, $Z_2\in\mathbb{R}^6$, $Z_e\in\mathbb{R}^6$, $\forall e\in\mathcal{E}$ are all sampled from standard normal. $Z_1$ and $Z_2$ are shared across environments and $Z_e$ is not. In each environment $e$, we sample 1,000 points. Again, data from original environments are combined, followed by the application of environment partitioning strategies (to obtain the same number of environments) and then IRM.
	In this experiment, we set $f$ as the identity function. After training a logistic classifier, we generate 5,000 different testing environments (each with a new $Z_e$ drawn from standard normal) to compute the worst-case prediction error rate. The penalty weight for IRM is $\lambda = 10^4$. Most hyper-parameters align with those in the study by \cite{rosenfeld2020risks}. The results, presented in Fig. \ref{RiskofIRM}, indicate that Decorr consistently yields a stable low error rate. This clearly demonstrates the superiority of Decorr.

	\subsection{Real Data Experiments}
	\label{detail}
	\subsubsection{Implementation Details}
	For each task, we utilize MLPs with two hidden layers featuring tanh activations and a dropout rate of $p=0.5$ following each hidden layer. The size of each hidden layer is $2^{\text{int}(\log_2 p)+2}$, where $p$ is the input dimension. The output layer is either linear or logistic. We employ Adam optimization \cite{kingma2014adam} with default parameters (learning rate $= 0.001$, $\beta_1 = 0.9$, $\beta_2 = 0.999$, $\varepsilon =  10^{-8}$) to minimize binary cross-entropy loss for classification and MSE for regression. The training involves a maximum of 20,000 iterations and includes an $L_2$ penalty term weighted by $0.001$. For IRM, the penalty weight is $\lambda = 10^4$. Except for occupancy estimation, all other experiments utilize $k = 2$ for Decorr and comparable methods, i.e., partitioning to create 2 environments. Unless otherwise specified, the original environment partitions for IRM and REx are determined based on the timestamps of the observations.
	
	\subsubsection{Financial Indicators}
	The task for the financial indicators dataset\footnote{https://www.kaggle.com/datasets/cnic92/200-financial-indicators-of-us-stocks-20142018} is to predict whether a stock's price will increase over the following year based on the stock's financial indicators. The dataset is divided into five annual segments from 2014 to 2018. Following the implementation details in \cite{krueger2021out}, we treat each year as a baseline environment, utilizing three environments for training, one for validation through early stopping, and one for testing. This setup results in a total of 20 different tasks. Additionally, we compile another set of 20 tasks, each using a single environment for training, one for validation, and three for testing. In this configuration, the original IRM and REx are ineffective due to the presence of only one explicit training environment. Following \cite{shen2021towards}, we assess all methods based on average error rate, worst-case error rate, and standard deviation across tasks in each task set. The results, displayed in Table \ref{Table:FI1}, indicate that Decorr achieves the best performance. The superiority of Decorr+IRM over pure IRM suggests that the original environment partitions, determined by the timestamps of observations, are suboptimal.
	
	\begin{table*}[t]
		\caption{The experiment on the Financial Indicators dataset. There are five pre-determined original environments. Task set 1 utilizes three environments for training, while task set 2 uses one environment for training. Each task set comprises 20 tasks, with the average error rate, worst-case error rate, and standard deviation across these tasks reported.}
		\label{Table:FI1}
		\begin{center}
				\begin{tabular}{ccccccc} 
					\toprule
					\multicolumn{1}{c}{} & \multicolumn{3}{c}{Task Set 1} & \multicolumn{3}{c}{Task Set 2} \\
					\cmidrule(r){2-4} \cmidrule(r){5-7} 
					Method & Average Error & Worst Error & STD & Average Error & Worst Error & STD \\
					\midrule 
					ERM & 45.02 (0.00) & 52.92 (0.12) & 5.20&  46.79 (0.01) & 51.05 (0.05) & \textbf{2.44}\\ 
					Random + IRM & 45.95 (0.28)  & 53.77 (0.58) & \textbf{4.27} & 46.57 (0.26)  & 52.03 (0.79) & 2.57\\
					EIIL  & 48.95 (0.16) & 57.37 (0.65) & 4.52& 48.97 (0.38) & 54.62 (0.70) & 2.71\\
					$k$-means + IRM & 45.20 (0.34) & 53.64 (0.44) & 6.19 & 46.79 (0.28) & 54.72 (1.22) & 3.43\\
					HRM  & 44.64 (0.00) & 53.52 (0.03) & 5.71 & 46.71 (0.00) & 51.17 (0.00) & 2.48\\
					\textbf{Decorr + IRM} & \textbf{43.99 (0.10)} & \textbf{51.61 (0.04)} & 5.33& \textbf{43.96 (0.25)} & \textbf{50.23 (1.17)} & 2.81\\
					\midrule
					IRM & 45.86 (0.14) & 56.38 (0.21) & 6.27\\
					V-REx & 44.08 (0.02) & 55.41 (0.12) & 6.30\\
					\bottomrule
				\end{tabular}
			\end{center}
		\end{table*}

		\subsubsection{Adult}
\begin{figure*}[t]
			\centering
			\begin{minipage}[t]{0.32\textwidth}
				\centering
				\includegraphics[width = \textwidth]{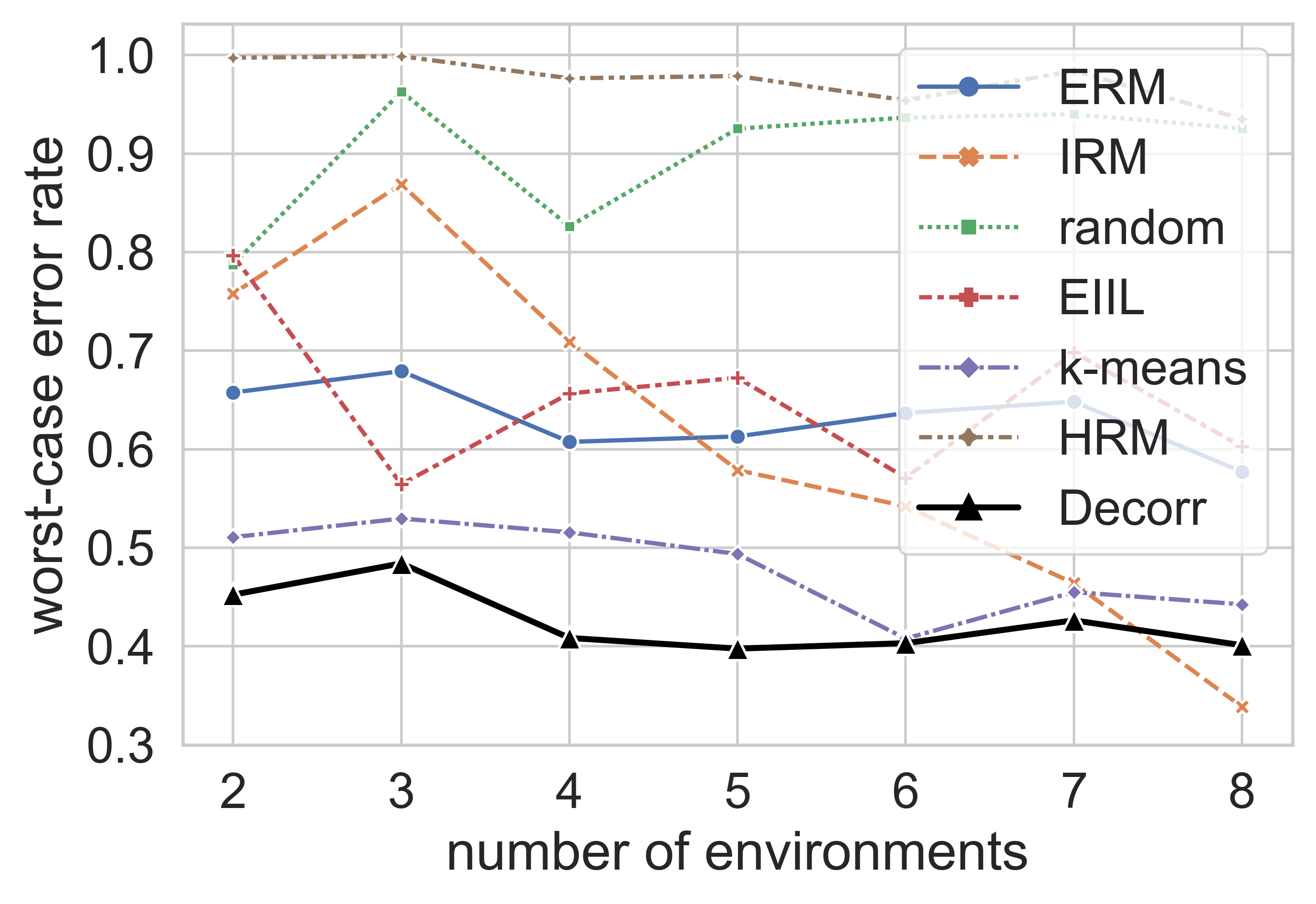}
				\caption{The example of Risks of IRM.}
				\label{RiskofIRM}
			\end{minipage}\hfill
			\begin{minipage}[t]{0.32\textwidth}
				\centering
				\includegraphics[width = \textwidth]{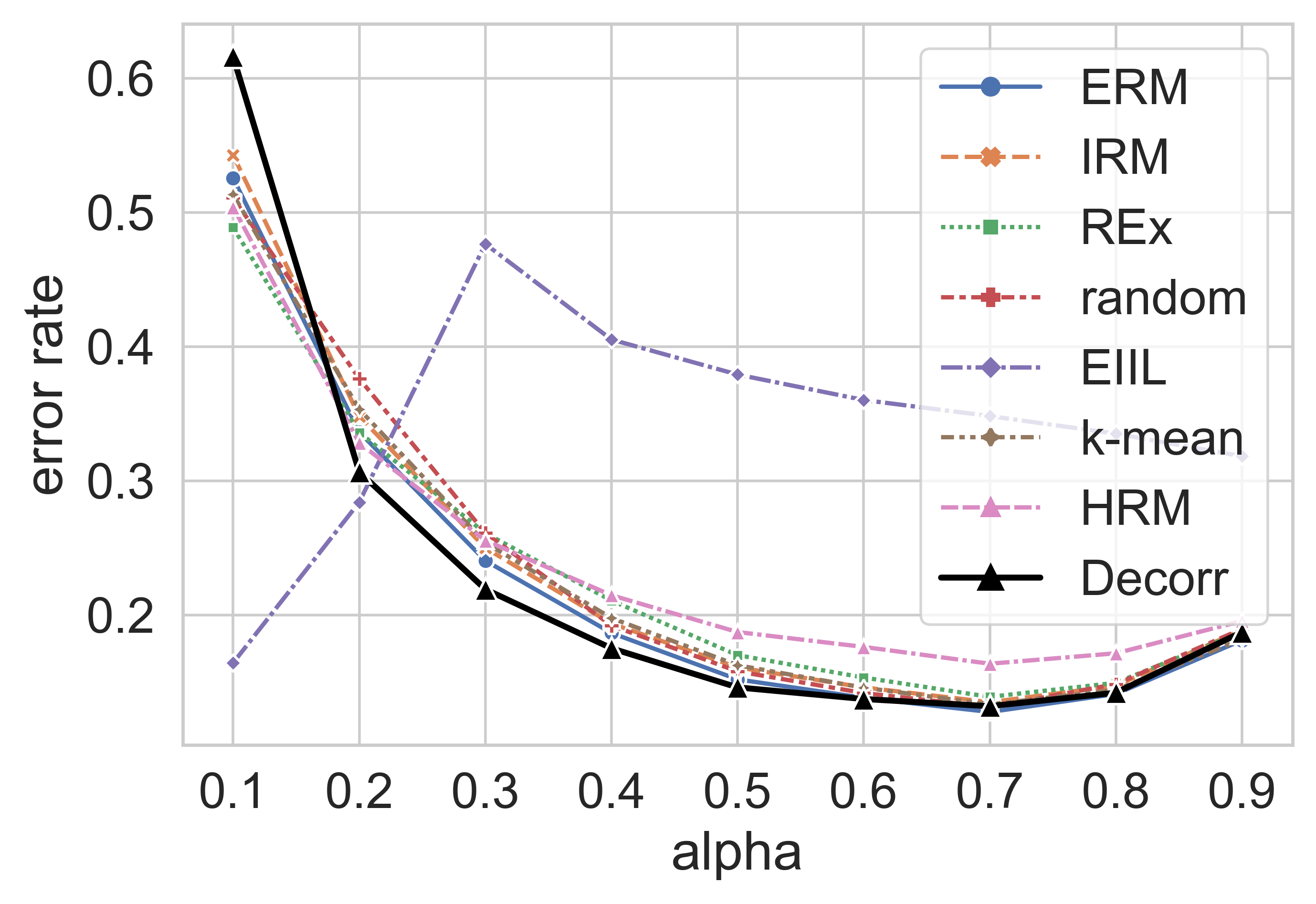}
				\caption{The Adult dataset: using race as the bias feature.}
				\label{adult1}
			\end{minipage}\hfill
			\begin{minipage}[t]{0.32\textwidth}
				\centering
				\includegraphics[width = \textwidth]{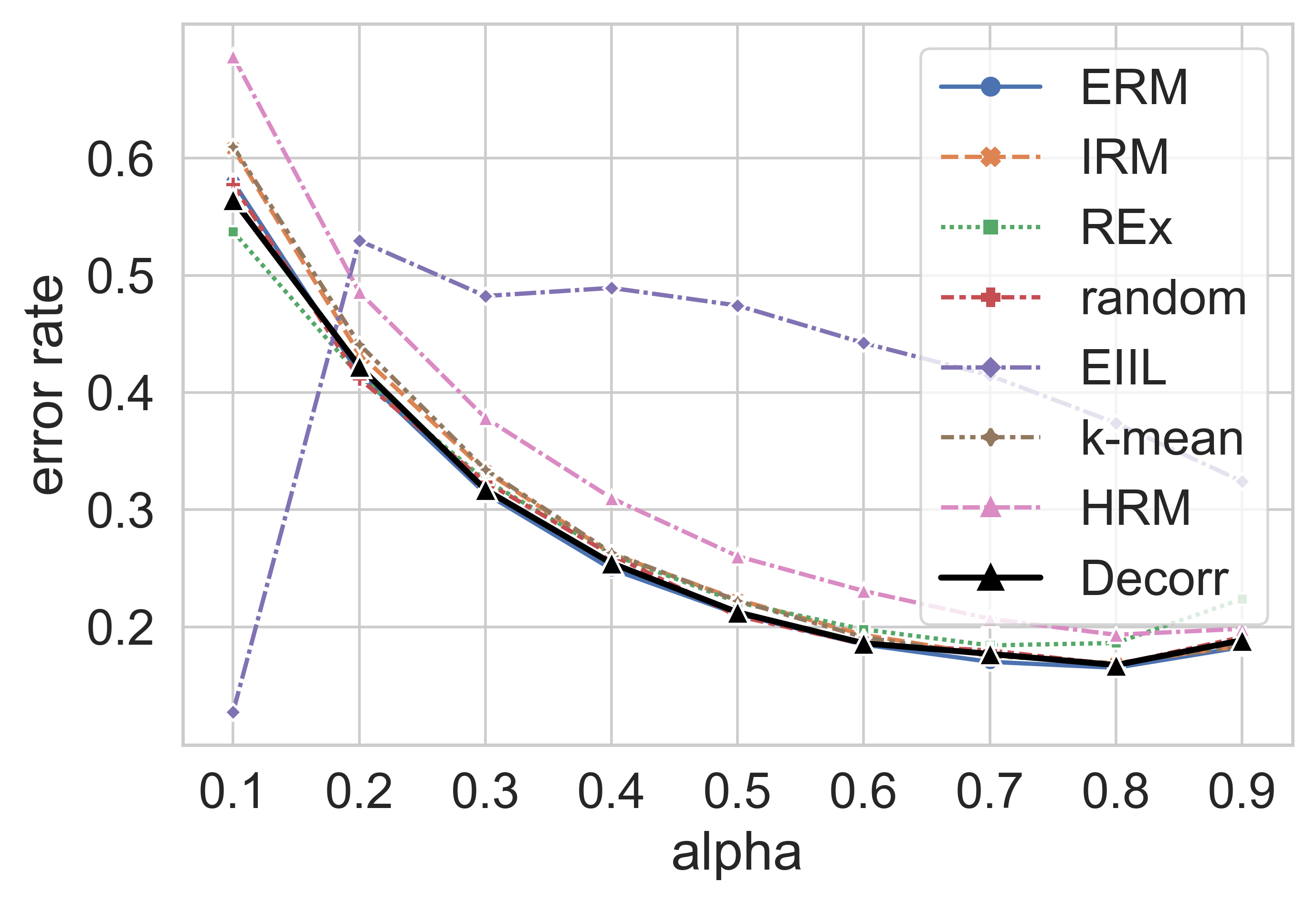}
				\caption{The Adult dataset: using sex as the bias feature.}
				\label{adult2}
			\end{minipage}  
		\end{figure*}

		The Adult dataset\footnote{https://archive.ics.uci.edu/ml/datasets/adult} is a tabular collection derived from a U.S. census, aimed at classifying whether an individual's annual income exceeds or falls below 50,000 USD based on specific characteristics. We retain only race and sex as categorical variables, converting them into binary values $\{0,1\}$. To introduce a distributional shift between training and testing data, we first identify either race or sex as the biased feature, denoted as $x^b$, and categorize the dataset into four groups: $\{x^b = 0, y = 0\}$, $\{x^b = 1, y = 1\}$, $\{x^b = 0, y = 1\}$, and $\{x^b = 1, y = 0\}$. For the training set, we use 90\% of the data from the first two groups and only an $\alpha$ proportion from the last two groups, with the remainder allocated to the testing set. A lower $\alpha$ value increases the distributional shift, whereas an $\alpha$ of 0.9 results in no shift. We anticipate that IRM-based methods will learn the spurious correlation ($x^b = 1$ leads to $y = 1$) under small $\alpha$ values. The error rate results on the testing set are depicted in Fig. \ref{adult1} and \ref{adult2}. For pure IRM and REx, original environments are defined by the value of the biased feature.
		
		\subsubsection{Occupancy Estimation}
		
		The Occupancy Estimation dataset\footnote{https://archive.ics.uci.edu/ml/datasets/Room+Occupancy+Estimation} comprises data from various sensors (temperature, light, sound, CO2, etc.) recorded every minute in a room. The task is to estimate the room occupancy, i.e., the count of individuals in the room. We treat this as a regression task, converting the time into a real number within the range $[0,1]$, standardizing the features, and training the models using the designated training and testing sets. The training errors are high for all the environment partitioning methods when $k = 2$, hence we divide to create $k = 3$ environments. The MSE on the testing set is displayed in Table \ref{occupancy}, where Decorr once again yields the lowest testing error. For pure IRM and REx, original environments are determined by the collection times of the samples.
		
		\begin{table}[t]
			\caption{The Occupancy Estimation dataset.}
			\label{occupancy}
				\begin{center}
					\begin{tabular}{cccc} 
						\toprule
						\multicolumn{1}{c}{Method}  &\multicolumn{1}{c}{Training Error} & \multicolumn{1}{c}{Testing Error}  \\
						\midrule
						ERM  & \textbf{0.026 (0.000)} & 0.773 (0.000)\\ 
						Random + IRM & 0.587 (0.053)  & 0.824 (0.093) \\
						EIIL  & 0.191 (0.203) & 0.690 (0.048) \\
						$k$-means + IRM & 0.044 (0.001) &  0.352 (0.009) \\
						HRM  & 0.166 (0.000) & 1.154 (0.000)\\
						\textbf{Decorr + IRM} & 0.143 (0.009) & \textbf{0.268 (0.011)} \\
						\midrule
						IRM & 0.420 (0.006) & 0.840 (0.021)   \\
						V-REx & 0.060 (0.001) & 0.686 (0.030)\\
						\bottomrule
					\end{tabular}
				\end{center}
			\end{table}
		
		\begin{table}[t]
			\caption{The Stock dataset.}
			\label{stock}
				\begin{center}
					\begin{tabular}{cccc} 
						\toprule
						\multicolumn{1}{c}{Method}  & \makecell{Average\\ Error} & \makecell{Worst-Case\\ Error} & \multicolumn{1}{c}{STD}  \\
						\midrule
						ERM  &  50.33 (0.10) & 56.65 (0.26) & 3.86\\ 
						Random + IRM & 50.11 (0.35)  & 54.54 (0.81) & 2.62\\
						EIIL  & 49.70 (0.33) & \textbf{52.29 (0.44)} & \textbf{1.65}\\
						$k$-means + IRM & 50.02 (0.30) & 54.92 (0.23) & 2.68\\
						HRM & 50.22 (0.00) & 55.97 (0.00) & 4.78\\
						\textbf{Decorr + IRM} & \textbf{49.13 (0.34)} & 53.50 (0.75) & 3.66 \\
						\bottomrule
					\end{tabular}
				\end{center}
			\end{table}
		
		\subsubsection{Stock}
		The Stock dataset\footnote{https://www.kaggle.com/datasets/nikhilkohli/us-stock-market-data-60-extracted-features} comprises market data and technical indicators for 10 U.S. stocks from 2005 to 2020. We attempt to predict whether the closing price of a stock will be higher tomorrow than it is today using today's technical indicators. For each stock, we allocate the first 70\% of the data as the training set, 10\% as the validation set for early stopping, and the final 20\% as the testing set. The modeling is conducted on a per-stock basis. We assess methods based on average error rate, worst-case error rate, and standard deviation across different stocks. Results are presented in Table \ref{stock}, where Decorr is shown to perform the best in terms of average error. Since there are no pre-defined environments, the original IRM and REx are not utilized.

			\subsection{Image Data Experiments}
			To demonstrate the applicability of our method across diverse data types, we implement Decorr and other baseline methods on two image datasets. With minor adjustments to Decorr tailored to the specifics of image data, our method consistently outperforms the others. Unless specified otherwise, the implementation details follow those outlined in Section \ref{detail}.
			
			Image data have much higher dimensions than tabular data, making direct decorrelation of raw image data impractical. We first need to extract features using a trained convolutional neural network, using the output from the last fully-connected layer as the data's features. With these extracted low-dimensional features, we can proceed with decorrelation as before (we apply $k$-means in this manner as well). After partitioning the dataset, we continue to use the raw image data for IRM training. For Decorr, we do not enforce uniform sample sizes across partitioned environments (i.e., $\lambda = 0$ in Algorithm \ref{algo1}), allowing for more diversified environments in image datasets.
			
			\subsubsection{Colored MNIST}
			Colored MNIST (CMNIST), introduced by \cite{arjovsky2019invariant} to assess IRM's capability to learn nonlinear invariant predictors, is an image dataset derived from MNIST. In CMNIST, each MNIST image is colored either red or green, creating a strong but spurious correlation between the image's color and its class label. This correlation poses a challenge for regular deep learning models, which tend to classify images based on color rather than shape.
			
\begin{figure}[t]
				\centering
				\begin{minipage}[t]{0.2\textwidth}
					\centering
					\includegraphics[width = \textwidth]{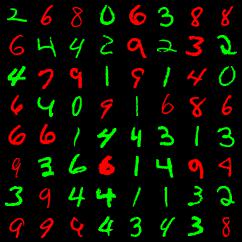}
					\caption{CMNIST training set examples: most of 0-4 are green, and most of 5-9 are red.}
					\label{CMNIST train}
				\end{minipage}\hspace{2em}
				\begin{minipage}[t]{0.2\textwidth}
					\centering
					\includegraphics[width = \textwidth]{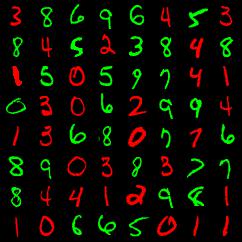}
					\caption{CMNIST testing set examples: most of 0-4 are red, and most of 5-9 are green.}
					\label{CMNIST test}
				\end{minipage}
			\end{figure}

			CMNIST is framed as a binary classification task, following the construction guidelines in \cite{arjovsky2019invariant}. Initially, images are assigned a preliminary label $\tilde{y}=0$ for digits 0-4 and $\tilde{y}=1$ for digits 5-9. The final label $y$ is then derived by flipping $\tilde{y}$ with a 0.1 probability, which introduces data noise. Subsequently, the color ID $z$ is sampled by flipping $y$ with a probability $p^e$, set at 0.2 for the first training environment and 0.1 for the second. In contrast, the test environment has a $p^e$ of 0.9, indicating a significant reversal in the correlation between color and label. The images are then colored according to their color ID $z$. Figures \ref{CMNIST train} and \ref{CMNIST test} provide a visual representation of this setup.
			
			Following \cite{arjovsky2019invariant}, we employ an MLP with two hidden layers as our base model. The network architecture and all hyperparameters remain consistent. We assess the methods by computing the error rate on the testing set. We also compute the percentage of data points exhibiting rare patterns (i.e., red images labeled 0-4 or green images labeled 5-9) within each partitioned environment. A significant discrepancy in percentages between two partitioned environments suggests greater diversity, which is potentially advantageous for invariant learning. Each model undergoes training for 5,000 epochs, with the initial 100 epochs proceeding without an IRM penalty. The outcomes, detailed in Table \ref{CMNIST}, demonstrate Decorr's great superiority.
			
			\begin{table}[t]
				\caption{The CMNIST experiment. Decorr achieves the lowest error rate on the testing set, compared to other baseline environment partitioning methods.}
				\label{CMNIST}
				\begin{center}
					\begin{tabular}{cccc} 
						\toprule
						\multicolumn{1}{c}{}  &\multicolumn{1}{c}{} &\multicolumn{2}{c}{\% of Rare Patterns} \\
						\cmidrule(r){3-4}
						Method & Average Error & Environment 1 & Environment 2\\
						\midrule
						ERM  & 43.33 (0.14)&--&--\\
						Random + IRM &  40.16 (0.16)&15.15 (0.08)&14.83 (0.10)\\
						EIIL  & 76.95 (3.46)&29.59 (2.14)&14.07 (0.09)\\
						$k$-means + IRM & 41.93 (0.52)&15.32 (0.21)&14.68 (0.19)\\
						\textbf{Decorr + IRM} & \textbf{34.89 (6.24)}& 40.24 (0.49) & 4.72 (0.26)\\
						\midrule
						IRM & 30.18 (0.94)&19.89 (0.13)&10.09 (0.12)\\
						V-REx & 34.22 (0.04)&19.89 (0.13)&10.09 (0.12)\\
						\bottomrule
					\end{tabular}
				\end{center}
			\end{table}
			
			\subsubsection{Waterbirds}
			The Waterbirds dataset, introduced by \cite{sagawa2019distributionally}, merges elements from the CUB dataset \cite{wah2011caltech} and the Places dataset \cite{zhou2017places}. The primary task is to identify the type of bird (\emph{waterbird} or \emph{landbird}) from the CUB dataset. However, combining CUB with Places introduces a spurious correlation: in the training set, most landbirds appear against land backgrounds and most waterbirds against water backgrounds, with 95\% of the data following this pattern. Conversely, in the testing set, only half of the data maintain this regular pattern, while the other half do not, thus the training set correlation does not hold. Further details about the dataset are available in \cite{sagawa2019distributionally}.
			
			Following \cite{sagawa2019distributionally}, we utilize a pretrained ResNet50 as our base model, setting the $L_2$ penalty weight at $10^{-4}$. Each model undergoes training for 50 epochs. We also apply IRM and V-REx using an optimal environment partition: one environment contains all regular pattern data (waterbirds in water and landbirds on land), and the other comprises all rare pattern data. We implement each method and record the error rate. The results, presented in Table \ref{Waterbirds}, show that Decorr achieves the best performance among all environment partitioning methods.
			
			\begin{table}[t]
				\caption{The Waterbirds experiment. Decorr achieves the lowest error rate on the testing set, compared to other baseline environment partitioning methods.}
				\label{Waterbirds}
				\begin{center}
					\begin{tabular}{cc} 
						\toprule
						\multicolumn{1}{c}{Method}  &\multicolumn{1}{c}{Average Error} \\
						\midrule
						ERM  & 25.76 (0.23)\\ 
						Random + IRM & 22.96 (0.43) \\
						EIIL  & 27.43 (7.25)\\
						$k$-means + IRM & 24.32 (0.65)\\
						\textbf{Decorr + IRM} & \textbf{22.70 (0.44)}\\
						\midrule
						IRM & 22.36 (0.08)\\
						V-REx & 33.53 (7.84)\\
						\bottomrule
					\end{tabular}
				\end{center}
			\end{table}

			\subsection{Summary}
			In this section, we briefly summarize the characteristics of the tested methods. IRM performs exceptionally well when the environment partition aligns with the true data-generating mechanisms. However, this alignment is often not present in real datasets. We observe that the simple and natural split-by-time partition offers no significant advantage. 
			$k$-means provides a straightforward approach to environment partitioning and consistently demonstrates good performance across multiple experiments. However, as illustrated in Fig. \ref{partition}, $k$-means may not always be the optimal method for partitioning. Our Decorr algorithm consistently achieves the best performance in the experiments mentioned above. Notably, Decorr does not aim to reconstruct the original data sources; rather, it seeks to identify partitions that exceed the performance of the original or natural partitions.
			
			\section{Conclusion}
			
			Invariant learning is a robust framework for Out-Of-Distribution (OOD) generalization, with environment partitioning playing a critical role in the effectiveness of IRM. Although existing partitioning methods perform well in some cases, their efficacy varies and they are not supported by a clear interpretation or justification. Inspired by the advantages of a low-correlated training set, we developed the Decorr algorithm, which partitions data into multiple environments with minimal internal correlation. We further provide explanations that uncorrelated environments enhance OOD generalization. Our partitioning approach offers distinct benefits over existing methods. Across various types of tasks, including image-based ones, we demonstrate that our method consistently and significantly enhances the performance of IRM, greatly broadening its applicability.




\bibliographystyle{IEEEtran}
\bibliography{ref}



\vfill

\end{document}